\pdfoutput=1
\documentclass[journal]{IEEEtran}

\usepackage{amsmath,graphicx}
\usepackage{booktabs}
\usepackage{multirow}
\usepackage{xcolor}
\usepackage{amsfonts}
\usepackage[inline]{enumitem}
\usepackage{mathtools}
\usepackage{amsthm}
\usepackage{amssymb}
\usepackage{pgfplots}
\usepackage{tikz}
\usetikzlibrary{arrows,automata,positioning}
\usepackage{caption}
\usepackage{subcaption}
\usepackage{relsize}
\usepackage{url}
\usepackage{hyperref}

\usepackage{cite}
\usepackage{url}
\usepackage{amsmath,amsfonts}
\usepackage{booktabs}
\usepackage{multirow, makecell}
\usepackage{footmisc}
\usepackage{pifont}
\usepackage{textcomp}
\usepackage{xcolor}

\pgfplotsset{compat=1.17}

\begin{document}
\title{HuBERT: Self-Supervised Speech Representation Learning by Masked Prediction of Hidden Units}
\author{Wei-Ning~Hsu,
        Benjamin~Bolte,
        Yao-Hung~Hubert~Tsai,
        Kushal Lakhotia,
        \\Ruslan~Salakhutdinov,
        Abdelrahman~Mohamed}

\maketitle

\begin{abstract}
Self-supervised approaches for speech representation learning are challenged by three unique problems: (1) there are multiple sound units in each input utterance, (2) there is no lexicon of input sound units during the pre-training phase, and (3) sound units have variable lengths with no explicit segmentation. To deal with these three problems, we propose the Hidden-Unit BERT (HuBERT) approach for self-supervised speech representation learning, which utilizes an offline clustering step to provide aligned target labels for a BERT-like prediction loss. A key ingredient of our approach is applying the prediction loss over the masked regions only, which forces the model to learn a combined acoustic and language model over the continuous inputs. HuBERT relies primarily on the consistency of the unsupervised clustering step rather than the intrinsic quality of the assigned cluster labels. Starting with a simple k-means teacher of 100 clusters, and using two iterations of clustering, the HuBERT model either matches or improves upon the state-of-the-art wav2vec 2.0 performance on the Librispeech (960h) and Libri-light (60,000h) benchmarks with 10min, 1h, 10h, 100h, and 960h fine-tuning subsets. Using a 1B parameter model, HuBERT shows up to 19\% and 13\% relative WER reduction on the more challenging dev-other and test-other evaluation subsets.\footnote{The code, pre-trained and fine-tuned models are available at \url{https://github.com/pytorch/fairseq/tree/master/examples/hubert}.}
\end{abstract}

\begin{IEEEkeywords}
Self-supervised learning, BERT.
\end{IEEEkeywords}

\IEEEpeerreviewmaketitle

\section{Introduction}
\label{sec:intro}
The north star for many research programs has been learning speech and audio representations through listening and interaction, similar to how babies learn their first language. High fidelity speech representation includes disentangled aspects of the spoken content along with non-lexical information of how it is delivered, e.g., speaker identity, emotion, hesitation, interruptions. Furthermore, reaching a complete situational understanding requires modeling structured noise interleaving and overlapping with the speech signal, e.g., laughter, coughing, lip-smacking, background vehicle engine, birds chirping, or food sizzling sounds.

The need for such high-fidelity representations drove research in self-supervised learning for speech and audio where the targets driving the learning process of a designed pretext task are drawn from the input signal itself. Examples of pretext tasks for self-supervised speech representation learning include distinguishing near-by features from temporally distant ones \cite{oord2018representation,schneider2019wav2vec,kharitonov2020data}, next-step prediction of audio features \cite{chung2019unsupervised}, masked prediction of audio features given unmasked context \cite{baevski2019vq,baevski2020wav2vec}. Besides, self-supervised learning methods do not rely on any linguistic resources during training, allowing them to learn universal representations since labels, annotations, and text-only material ignores rich information in the input signal.

Learning speech representations without reliance on large volumes of labeled data is crucial for industrial applications and products with ever-increasing coverage of new languages and domains. The time needed to collect large labeled datasets covering each of these scenarios is the real bottleneck in the current fast-moving AI industry, with time-to-market playing a critical role for product success. Building more inclusive applications covering spoken-only dialects and languages is another significant benefit of reducing dependence on linguistic resources. Given their non-standard orthographic rules, many of these languages and dialects have very little or no resources at all.

Pseudo-labeling (PL), also known as self-training and belongs to the family of semi-supervised learning techniques, has been the dominant approach for utilizing unlabeled speech and audio with successful applications dating back to the mid-1990s \cite{Zavaliagkos_98, ma_bbn_06, kahn2020self, hsu2020semi}. PL starts with some supervised data to train a "teacher" model in one specific downstream task. Pseudo-labels are then generated for the unlabeled data using the teacher model. Next, a student model is trained using the combined supervised and teacher-labeled data either using the standard cross-entropy \cite{kahn2020self} loss or using a contrastive loss \cite{xiao2021contrastive} to account for noise in teacher-generated labels. The pseudo-labeling process may be repeated multiple times to improve teacher label quality \cite{xu2020iterative} iteratively.

Without discounting the immense success of pseudo-labeling techniques, self-supervised representations offer two unique advantages: (1) Pseudo-label methods force student models to merely mimic a teacher model, which is limited by its supervised data size and the provided annotation quality. On the other hand, self-supervised pretext tasks force the model to represent the entire input signal by compressing much more bits of information into the learned latent representation. (2) In pseudo-labeling, the supervised data of the teacher model forces the whole learning to be geared towards a single downstream task. On the contrary, self-supervised features show better generalization to a multitude of downstream applications.

There have been impressive successes for self-supervised learning in Computer Vision (CV) \cite{caron2020Swav, Chen2020SimSiam, grill2020byol} and Natural Language Processing (NLP) \cite{brown2020gpt3, liu2019roberta, lewis2019bart} applications. Learning representations of discrete input sequences, such as in Natural Language Processing (NLP) applications, uses either masked prediction \cite{devlin2018bert, clark2020electra} or auto-regressive generation \cite{peters2018deep, lewis2019bart} of input sequences with partial obfuscation. For continuous inputs, such as in Computer Vision (CV) applications, representations are often learned through instance classification, in which each image and its augmentations are treated as a single output class to be pulled together \cite{Chen2020SimSiam, grill2020byol} or contrasted against other negative samples \cite{he2020momentum}.

Speech signals differ from text and images in that they are \textit{continuous-valued} \textit{sequences}. Self-supervised learning for the speech recognition domain faces unique challenges from those in CV and NLP. Firstly, the presence of multiple sounds in each input utterance breaks the instance classification assumption used in many CV pre-training approaches. Secondly, during pre-training, there is no prior lexicon of discrete sound units available, as in NLP applications in which words or word pieces are used, hindering the use of predictive losses. Lastly, the boundaries between sound units are not known, which complicates masked prediction pre-training.

In this paper, we introduce \textbf{H}idden \textbf{u}nit \textbf{BERT} (HuBERT) that benefits from an offline clustering step to generate noisy labels for a BERT-like per-training. Concretely, a BERT model consumes masked continuous speech features to predict pre-determined cluster assignments. The predictive loss is only applied over the masked regions, forcing the model to learn good high-level representations of unmasked inputs to infer the targets of masked ones correctly. Intuitively, the HuBERT model is forced to learn both acoustic and language models from continuous inputs. First, the model needs to model unmasked inputs into meaningful continuous latent representations, which maps to the classical acoustic modeling problem. Second, to reduce the prediction error, the model needs to capture the long-range temporal relations between learned representations. One crucial insight motivating this work is the importance of consistency of the targets, not just their correctness, which enables the model to focus on modeling the sequential structure of input data. Our approach draws inspiration from the DeepCluster method for self-supervised visual learning \cite{caron2018deep}; however, HuBERT benefits from the masked prediction loss over speech sequences to represent their sequential structure.

When the HuBERT model is pre-trained on either the standard Librispeech 960h \cite{panayotov2015librispeech} or the Libri-Light 60k hours \cite{kahn2020libri}, it either matches or improves upon the state-of-the-art wav2vec 2.0 \cite{baevski2020wav2vec} performance on all fine-tuning subsets of 10mins, 1h, 10h, 100h, and 960h. We present systematic results on three model sizes pre-trained with HuBERT: \textsc{Base} (90M parameters), \textsc{Large} (300M), and \textsc{X-Large} (1B). The \textsc{X-Large} model shows up to 19\% and 13\% relative WER improvement from \textsc{Large} models on dev-other and test-other evaluation subsets when pre-trained on the Libri-Light 60k hours.
\section{Method}
\subsection{Learning the Hidden Units for HuBERT}
An acoustic model trained on text and speech pairs provides pseudo-phonetic labels for each frame via forced alignment in semi-supervised learning. On the contrary, the self-supervised representation learning setup has access to speech-only data. Nevertheless, simple discrete latent variable models such as k-means and Gaussian mixture models (GMMs) infer hidden units that exhibit non-trivial correlation with the underlying acoustic units~\cite{lee2012nonparametric} (see also Table~\ref{tab:loss}). More advanced systems can achieve better acoustic unit discovery performance using better graphical models \cite{ondel2016variational, ebbers2017hidden} or parameterizes the distributions with more powerful neural network models~\cite{hsu2017learning, hsu2017unsupervised, chorowski2019unsupervised, khurana2019factorial, khurana2020convolutional}.

\begin{figure}[th]
  \centering
  \includegraphics[width=\linewidth]{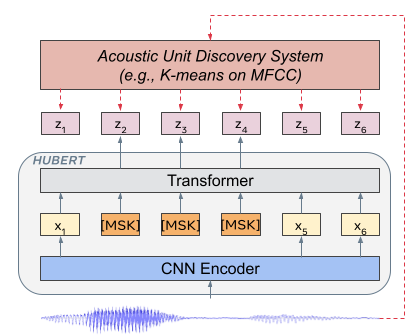}\caption{The HuBERT approach predicts hidden cluster assignments of the masked frames ($y_2, y_3, y_4$ in the figure) generated by one or more iterations of k-means clustering.}
  \label{fig:arch}
\end{figure}
Inspired by this, we propose to use acoustic unit discovery models to provide frame-level targets. 
Let $X$ denote a speech utterance $X = [x_1, \cdots, x_T]$ of $T$ frames.
Discovered hidden units are denoted with $h(X) = Z = [z_1, \cdots, z_T]$, where $z_t \in [C]$ is a $C$-class categorical variable and $h$ is a clustering model, e.g. k-means. 

\subsection{Representation Learning via Masked Prediction}\label{sec:maskpred}
Let $M \subset [T]$ denote the set of indices to be masked for a length-$T$ sequence $X$, and $\tilde{X} = r(X, M)$ denote a corrupted version of $X$ where $x_t$ is replaced with a mask embedding $\tilde{x}$ if $t \in M$. A masked prediction model $f$ takes as input $\tilde{X}$ and predicts a distribution over the target indeces at each timestep $p_f(\cdot \mid \tilde{X}, t)$. There are two decisions to be made for masked prediction: \textit{how to mask} and \textit{where to apply the prediction loss}. 

Regarding the first decision, we adopt the same strategies used in SpanBERT~\cite{joshi2020spanbert} and wav2vec 2.0~\cite{baevski2020wav2vec} for mask generation, where $p$\% of the timesteps are randomly selected as start indices, and spans of $l$ steps are masked. To address the second decision, we denote the cross-entropy loss computed over masked and unmasked timesteps as $L_m$ and $L_u$, respectively. $L_m$ is defined as:
\begin{equation}
    L_m(f; X, M, Z) = \sum_{t \in M} \log p_f(z_t \mid \tilde{X}, t),
\end{equation}
and $L_u$ is of the same form except that it sums over $t \not\in M$.
The final loss is computed as a weighted sum of the two terms: $L = \alpha L_m + (1-\alpha)L_u$. In the extreme case when $\alpha = 0$, the loss is computed over the unmasked timesteps, which is similar to acoustic modeling in hybrid speech recognition systems \cite{young1996large, abdel2012applying, povey2005discriminative, bourlard2012connectionist}. In our setup, this limits the learning process to mimicking the clustering model. 

In the other extreme with $\alpha=1$, the loss is only computed over the masked timesteps where the model has to predict the targets corresponding to the unseen frames from context, analogous to language modeling. It forces the model to learn both the acoustic representation of unmasked segments and the long-range temporal structure of the speech data. We hypothesize that the setup with $\alpha=1$ is more resilient to the quality of cluster targets, which is demonstrated in our experiments (see Table~\ref{tab:loss}).

\subsection{Learning with Cluster Ensembles}
A simple idea to improve target quality is to utilize multiple clustering models. While an individual clustering model may perform terribly, cluster ensembles can provide complementary information to facilitate representation learning. For example, an ensemble of k-means models with different codebook sizes can create targets of different granularity, from manner classes (vowel/consonant) to sub-phone states (senones). 
To extend the proposed framework, let $Z^{(k)}$ be the target sequences generated by the $k$-th clustering model. We can now re-write $L_m$ as:
\begin{equation}
    L_m(f; X, \{ Z^{(k)} \}_k, M) = 
    \sum_{t \in M} \sum_{k} \log p_f^{(k)}(z_t^{(k)} \mid \tilde{X}, t), \label{eq:our_obj}
\end{equation}
and similarly for the unmasked loss $L_u$. This is analogous to multi-task learning, but with tasks created by unsupervised clustering.

Additionally, ensembling is intriguing because it can be used alongside product quantization (PQ)~\cite{gray1998quantization}, where a feature space is partitioned into multiple subspaces, and each subspace is quantized separately. PQ allows effective Euclidean distance-based quantization such as k-means for high-dimensional features and heterogeneous features whose scale differs significantly between subspaces. In this case, the theoretical size of the target space is the product of all codebooks' sizes.

\subsection{Iterative Refinement of Cluster Assignments}
In addition to using cluster ensembles, another direction for improved representation is \textit{refining} the cluster assignments throughout the learning process. Since we expect a pre-trained model to provide better representations than the raw acoustic feature such as MFCCs, we can create a new generation of clusters by training a discrete latent model over the learned latent representations. The learning process then proceeds with the newly discovered units.

\subsection{Implementation}\label{sec:impl}
Our pre-trained models follows the wav2vec 2.0 architecture~\cite{baevski2020wav2vec}, with a convolutional waveform encoder, a BERT encoder~\cite{devlin2018bert}, a projection layer and a code embedding layer. We consider HuBERT in three different configurations: \textsc{Base}, \textsc{Large}, and \textsc{X-Large}. The fisrt two follow the architectures of wav2vec 2.0 \textsc{Base} and \textsc{Large} closely. The \textsc{X-Large} architecture expands the model size to about 1 billion parameters, similar to the size of the Conformer XXL model in ~\cite{zhang2020pushing}. 
The waveform encoder is identical for all the three configurations, which is composed of seven 512-channel layers with strides [5,2,2,2,2,2,2] and kernel widths [10,3,3,3,3,2,2]. The BERT encoder consists of many identical transformer blocks, whose parameters along with the parameter of the subsequent projection layer are specified in Table~\ref{tab:arch}.

\begin{table}[ht]
    \centering
    \begin{tabular}{cc|ccc}
        \toprule
         & & \textsc{Base} & \textsc{Large} & \textsc{X-Large} \\
        \midrule
        \multirow{3}{*}{CNN Encoder}
        & strides      & \multicolumn{3}{c}{5, 2, 2, 2, 2, 2, 2} \\
        & kernel width & \multicolumn{3}{c}{10, 3, 3, 3, 3, 2, 2} \\
        & channel      & \multicolumn{3}{c}{512} \\
        \midrule
        \multirow{5}{*}{Transformer} 
        & layer        & 12 & 24 & 48 \\
        & embedding dim. & 768 & 1024 & 1280 \\
        & inner FFN dim. & 3072 & 4096 & 5120 \\
        & layerdrop prob  & 0.05 & 0 & 0 \\
        & attention heads & 8 & 16 & 16 \\
        \midrule
        \multirow{1}{*}{Projection} & dim. & 256 & 768 & 1024 \\
        \midrule
        \multicolumn{2}{c|}{Num. of Params} & 95M & 317M & 964M \\
        \bottomrule
        
    \end{tabular}
    \caption{Model architecture summary for \textsc{Base}, \textsc{Large}, and \textsc{X-Large} HuBERT models}
    \label{tab:arch}
\end{table}

The convolutional waveform encoder generates a feature sequence at a 20ms framerate for audio sampled at 16kHz (CNN encoder down-sampling factor is 320x). The audio encoded features are then randomly masked as described in Section \ref{sec:maskpred}. The BERT encoder takes as input the masked sequence and outputs a feature sequence $[o_1, \cdots, o_T]$. The distribution over codewords is parameterized with
\begin{equation}
    p_f^{(k)}(c \mid \tilde{X}, t) = \frac{\exp(\text{sim}(A^{(k)} o_t, e_c) / \tau)} {\sum_{c'=1}^C \exp(\text{sim}(A^{(k)} o_t, e_{c'}) / \tau)},
\end{equation}
where $A$ is the projection matrix, $e_c$ is the embedding for codeword $c$, $\text{sim}(\cdot, \cdot)$ computes the cosine similarity between two vectors, and $\tau$ scales the logit, which is set to 0.1. When cluster ensembles are used, one projection matrix $A^{(k)}$ is applied for each clustering model $k$.

After HuBERT pre-training, We use the connectionist temporal classification (CTC)~\cite{graves2006connectionist} loss for ASR fine-tuning of the whole model weights except the convolutional audio encoder, which remains frozen. The projection layer(s) is removed and replaced with a randomly initialized softmax layer. The CTC target vocabulary includes 26 English characters, a space token, an apostrophe, and a special CTC blank symbol.

\section{Related Work}
We discuss recent studies on self-supervised speech representation learning by grouping them by training objective. The earliest line of work learns representations by postulating a generative model for speech with latent variables, which are assumed to capture the relevant phonetic information. Training of these models amounts to likelihood maximization. Different latent structures have been applied to encode the prior assumption, such as continuous~\cite{hsu2017learning}, discrete~\cite{chorowski2019unsupervised,van2017neural}, or sequential~\cite{hsu2017unsupervised,ebbers2017hidden,glarner2018full,khurana2019factorial,khurana2020convolutional}.

Prediction-based self-supervised learning has gathered increasing interests recently, where a model is tasked to predict the content of the unseen regions \cite{chung2019unsupervised, chung2020generative, chung2020improved, ling2020deep, wang2020unsupervised, liu2020mockingjay, chi2020audio, ling2020decoar} or to contrast the target unseen frame with randomly sampled ones \cite{oord2018representation, kharitonov2020data, schneider2019wav2vec, baevski2020wav2vec}. Some models combine both the predictive and the contrastive losses \cite{baevski2019vq, baevski2019effectiveness}. These objectives can usually be interpreted as mutual information maximization~\cite{tsai2020ssl_multi}. Other objectives do not belong to these categories, for example, \cite{pascual2019learning}.

This work is most related to DiscreteBERT~\cite{baevski2019effectiveness}: both HuBERT and DiscreteBERT predict discrete targets of masked regions. However, there are several crucial differences. First, instead of taking quantized units as input, HuBERT takes raw waveforms as input to pass as much information as possible to the transformer layers, which was shown to be important in \cite{baevski2020wav2vec}. Furthermore, in the experiment section, we show that our model, with simple k-means targets, can achieve better performance than DiscreteBERT that uses vq-wav2vec \cite{baevski2019vq} learned units. Second, we also present many techniques to improve teacher quality instead of using a single fixed teacher as done in DiscreteBERT.

HuBERT is also related to wav2vec 2.0~\cite{baevski2020wav2vec}. However, the latter employs a contrastive loss that requires careful design of where to sample negative frames from, an auxiliary diversity loss to encourage the discrete unit usage, and demands a proper Gumbel-softmax temperature annealing schedule. In addition, it only explores quantizing the waveform encoder output, which may not be the best feature for quantization due to the limited capacity of the convolutional encoder, as suggested by our ablation studies in Figure~\ref{fig:qual_layer}. Concretely, our proposed method adopts a more direct predictive loss by separating the acoustic unit discovery step from the masked prediction representation learning phase and achieves the state-of-the-art results that match or outperform wav2vec 2.0 on different fine-tuning scales.

Finally, the idea of iterative refinement target labels is similar to iterative pseudo labeling for semi-supervised ASR~\cite{xu2020iterative, likhomanenko2020slimipl}, which leverages an improving student model to generate better pseudo-labels for the next iteration of training. The HuBERT approach can be seen as extending this method to the self-supervised setup with a masked prediction loss.
\section{Experimental Details}

\subsection{Data}\label{sec:data}
For unsupervised pre-training, we use the full 960 hours of LibriSpeech audio~\cite{panayotov2015librispeech} or 60,000 hours of Libri-light~\cite{kahn2020libri} audio, both of which are derived from the LibriVox project that contains English recordings of copyright-free audiobooks by volunteers from the Internet.
For supervised fine-tuning, five different partitions are considered: Libri-light 10-minute, 1-hour, 10-hour splits and LibriSpeech 100-hour (\texttt{train-clean-100}) and 960-hour (\texttt{train-clean-100}, \texttt{train-clean-360}, \texttt{train-other-500} combined) splits. The three Libri-light splits are subsets of the the LibriSpeech training split, and each of them contain half of the audio from \texttt{train-clean-*} and the other from \texttt{train-other-500}.

\subsection{Unsupervised Unit Discovery}
To demonstrate the effectiveness of the proposed method on utilizing low-quality cluster assignments, we consider the k-means algorithm~\cite{lloyd1982least} for acoustic unit discovery by default. It is one of the most naive unit discovery models that can be treated as modeling an isotropic Gaussian with the same scalar variance for each acoustic unit.
To generate labels for the first iteration HuBERT training over the 960 hour LibriSpeech training set, we run k-means clustering with 100 clusters on 39-dimensional MFCC features, which are 13 coefficients with the first and the second-order derivatives.

To generate better targets for the subsequent iterations, we run k-means clustering with 500 clusters on the latent features extracted from the HuBERT model pre-trained in the previous iteration (not fine-tuned) at some intermediate transformer layer.
Since the feature dimension at the transformer output is much higher than the MFCC features (768-D for HuBERT \textsc{Base}), we cannot afford to load the entire 960 hour training split to the memory. So instead, we randomly sample 10\% of the data for fitting the k-means model.

The \texttt{MiniBatchKMeans} algorithm implemented in the \texttt{scikit-learn} \cite{pedregosa2011scikit} package is used for clustering, which fits a mini-batch of samples at a time.\footnote{It still requires loading the entire dataset to the memory first.} We set the mini-batch size to be 10,000 frames. k-means++~\cite{arthur2006k} with 20 random starts is used for better initialization.

\subsection{Pre-Training}\label{sec:pretrain}
We train the \textsc{Base} model for two iterations on the 960 hours of LibriSpeech audio on 32 GPUs, with a batch size of at most 87.5 seconds of audio per GPU. The first iteration is trained for 250k steps, while the second iteration is trained for 400k steps using labels generated by clustering the 6-th transformer layer output of the first iteration model. Training for 100k steps takes about 9.5 hours.

Next we train HuBERT \textsc{Large} and \textsc{X-Large} for one iteration on 60,000 hours of Libri-light audio on 128 and 256 GPUs, respectively, for 400k steps. The batch sizes are reduced to 56.25 and 22.5 seconds of audio per GPU due to memory constraints.
Instead of restarting the iterative process from clustering MFCC features, we extract features from the 9-th transformer layer of the second iteration \textsc{Base} HuBERT for clustering and use those labels for training these two models. Hence, these two models can also be seen as the third iteration models.

For all HuBERT configurations, mask span is set to $l=10$, and $p=8\%$ of the waveform encoder output frames are randomly selected as mask start if not otherwise mentioned. Adam~\cite{kingma2014adam} optimizer is used with $\beta = (0.9, 0.98)$, and the learning rate ramps up linearly from 0 to the peak learning rate for the first 8\% of the training steps, and then decays linearly back to zero. The peak learning rates are 5e-4/1.5e-3/3e-3 for \textsc{Base}/\textsc{Large}/\textsc{X-Large} models.

\subsection{Supervised Fine-Tuning and Decoding}
We fine-tune each model on 8 GPUs on the labeled splits described in Section~\ref{sec:data}. The batch sizes per GPU are at most 200/80/40 seconds of audio for \textsc{Base}/\textsc{Large}/\textsc{X-Large} models. During fine-tuning, the convolutional waveform audio encoder parameters are fixed. Like wav2vec 2.0, we introduce a \textit{freeze-step} hyperparameter to control how many fine-tuning steps the transformer parameters are fixed, and only the new softmax matrix is trained. 
We sweep over peak learning rate ([1e-5, 1e-4]), learning rate schedule (percentage of steps for linear ramp-up and decay), number of fine-tuning steps, freeze step, and waveform encoder output masking probability for each model size and fine-tuning split combination using the word error rate (WER) on the \texttt{dev-other} subset as a criterion for model selection.

We use the wav2letter++~\cite{pratap2018wav2letter++} beam search decoder wrapped in Fairseq~\cite{ott2019fairseq} for language model-fused decoding, which optimizes:
\begin{equation}
    \log p_{CTC}(Y \mid X) + w_1 \log P_{LM}(Y) + w_2 |Y|,
\end{equation}
where $Y$ is the predicted text, $|Y|$ is the length of the text, and $w_1$ and $w_2$ denote the language model weight and word score. The decoding hyperparameters are searched with Ax, a Bayesian optimization toolkit,\footnote{\url{https://github.com/facebook/Ax}}. In this work, we consider both $n$-gram and transformer language models trained on the official Librispeech language modeling data.

\subsection{Metrics of Target Quality}
For analysis, we derive frame-level forced-aligned phonetic transcripts using a hybrid ASR system to measure the correlation between the k-means cluster assignments and the actual phonetic units. 
Given aligned frame-level phonetic labels $[y_1, \cdots, y_T]$ and k-means labels $[z_1, \cdots, z_T]$, the joint distribution between the two variables $p_{yz}(i, j)$ can be estimated by counting the occurrences: 
\begin{equation}
    p_{yz}(i, j) = \dfrac{\sum_{t=1}^T [y_t = i \wedge z_t = j] }{T},
\end{equation}
where $i$ denotes the $i$-th phoneme class and $j$ denotes the $j$-th k-means label class.
The marginal probabilities are computed as $p_z(j) = \sum_i p_{yz}(i, j)$ and $p_y(j) = \sum_j p_{yz}(i, j)$.

For each phone class $i$, we further compute the most likely target label as:
\begin{equation}
    z^*(i) = \arg\max_j p_{yz}(i, j).
\end{equation}
Likewise, for each k-means class $j$, we compute the most likely phone label as:
\begin{equation}
    y^*(j) = \arg\max_i p_{yz}(i, j).
\end{equation}
Three metrics are considered:
\begin{enumerate}
    \item \textbf{phone purity} (Phn Pur.): 
    \begin{equation}
        \mathbb{E}_{p_z(j)} [ p_{y \mid z}(y^*(j) \mid j) ],
    \end{equation}
    where $p_{y \mid z}(i \mid j) = p_{yz}(i, j) / p_z(j)$ denotes the conditional probability of phone given a k-means label. This metric measures the average phone purity within one class, which can be interpreted as the frame-level phone accuracy if we transcribe each k-means class with its most likely phone label. When comparing different sets of target labels with the same number of units, higher purity indicates better quality. However, this metric is less meaningful when comparing two sets with different numbers of units: in the extreme case where each frame is assigned a unique target label, the phone purity would be 100\%.
    \item \textbf{cluster purity} (Cls Pur.):
    \begin{equation}
        \mathbb{E}_{p_y(i)} [ p_{z \mid y}(z^*(i) \mid i) ],
    \end{equation}
    where $p_{z \mid y}(j \mid i) = p_{yz}(i, j) / p_y(i)$ denotes the conditional probability of a k-means label given phone label. Cluster purity is the counterpart of phone purity, whose value would typically decrease when the number of units increases. When comparing target labels with the same number of units, higher cluster purity also indicates a better quality, as frames of the same phone are more likely labeled as the same k-means label class.
    \item \textbf{phone-normalized mutual information} (PNMI):
    \begin{align}
        \dfrac{I(y; z)}{H(y)} &= \dfrac{ 
            \sum_i \sum_j p_{yz}(i, j) \log \dfrac{p_{yz}(i, j)}{p_{y}(i)p_{z}(j)}
        }{
            \sum_i p_{y}(i) \log p_{y}(i)
        } \\
        &= \dfrac{H(y) - H(y \mid z)} {H(y)} \\
        &= 1 - \dfrac{H(y \mid z)} {H(y)}.
    \end{align}
    PNMI is an information-theoretic metric that measures the percentage of uncertainty about the phone label $y$ eliminated after observing the k-means label $z$. Higher PNMI also indicates better k-means clustering quality.
\end{enumerate}

\begin{table*}[t]
    \centering
    \begin{tabular}{lcccccc}
        \toprule
        Model & Unlabeled Data & LM & dev-clean & dev-other & test-clean & test-other \\
        \midrule\midrule
        \multicolumn{7}{c}{\textit{\textbf{10-min labeled}}} \\
        DiscreteBERT~\cite{baevski2019effectiveness} & LS-960 & 4-gram & 15.7 & 24.1 & 16.3 & 25.2 \\
        wav2vec 2.0 \textsc{Base}~\cite{baevski2020wav2vec}   & LS-960 & 4-gram & 8.9 & 15.7 & 9.1 & 15.6 \\
        wav2vec 2.0 \textsc{Large}~\cite{baevski2020wav2vec}  & LL-60k & 4-gram & 6.3 & 9.8 & 6.6 & 10.3  \\
        wav2vec 2.0 \textsc{Large}~\cite{baevski2020wav2vec}  & LL-60k & Transformer & 4.6 & 7.9 & 4.8 & 8.2 \\
        \midrule
        HUBERT \textsc{Base}    & LS-960 & 4-gram & 9.1 & 15.0 & 9.7 & 15.3  \\
        HUBERT \textsc{Large}   & LL-60k & 4-gram & 6.1 & 9.4 & 6.6 & 10.1 \\
        HUBERT \textsc{Large}   & LL-60k & Transformer & 4.3 & 7.0 & 4.7 & 7.6 \\
        HUBERT \textsc{X-Large} & LL-60k & Transformer & 4.4 & 6.1 & 4.6 & 6.8 \\
        
        \midrule\midrule
        \multicolumn{7}{c}{\textit{\textbf{1-hour labeled}}} \\
        DeCoAR 2.0~\cite{ling2020decoar} & LS-960 & 4-gram & - & - & 13.8 & 29.1 \\
        DiscreteBERT~\cite{baevski2019effectiveness} & LS-960 & 4-gram & 8.5 & 16.4 & 9.0 & 17.6 \\
        wav2vec 2.0 \textsc{Base}~\cite{baevski2020wav2vec}   & LS-960 & 4-gram & 5.0 & 10.8 & 5.5 & 11.3 \\
        wav2vec 2.0 \textsc{Large}~\cite{baevski2020wav2vec}  & LL-60k & Transformer & 2.9 & 5.4 & 2.9 & 5.8 \\
        \midrule
        HUBERT \textsc{Base}    & LS-960 & 4-gram & 5.6 & 10.9 & 6.1 & 11.3  \\
        HUBERT \textsc{Large}   & LL-60k & Transformer & 2.6 & 4.9 & 2.9 & 5.4 \\
        HUBERT \textsc{X-Large} & LL-60k & Transformer & 2.6 & 4.2 & 2.8 & 4.8 \\
        
        \midrule\midrule
        \multicolumn{7}{c}{\textit{\textbf{10-hour labeled}}} \\
        SlimIPL~\cite{likhomanenko2020slimipl} & LS-960 & 4-gram + Transformer & 5.3 & 7.9 & 5.5 & 9.0 \\
        DeCoAR 2.0~\cite{ling2020decoar} & LS-960 & 4-gram & - & - & 5.4 & 13.3 \\
        DiscreteBERT~\cite{baevski2019effectiveness} & LS-960 & 4-gram & 5.3 & 13.2 & 5.9 & 14.1 \\
        wav2vec 2.0 \textsc{Base}~\cite{baevski2020wav2vec}   & LS-960 & 4-gram & 3.8 & 9.1 & 4.3 & 9.5 \\
        wav2vec 2.0 \textsc{Large}~\cite{baevski2020wav2vec}  & LL-60k & Transformer & 2.4 & 4.8 & 2.6 & 4.9 \\
        \midrule
        HUBERT \textsc{Base}    & LS-960 & 4-gram & 3.9 & 9.0 & 4.3 & 9.4 \\
        HUBERT \textsc{Large}   & LL-60k & Transformer & 2.2 & 4.3 & 2.4 & 4.6 \\
        HUBERT \textsc{X-Large} & LL-60k & Transformer & 2.1 & 3.6 & 2.3 & 4.0 \\
        
        \midrule\midrule
        \multicolumn{7}{c}{\textit{\textbf{100-hour labeled}}} \\
        IPL~\cite{xu2020iterative} & LL-60k & 4-gram + Transformer & 3.19 & 6.14 & 3.72 & 7.11 \\
        SlimIPL~\cite{likhomanenko2020slimipl} & LS-860 & 4-gram + Transformer & 2.2 & 4.6 & 2.7 & 5.2 \\
        Noisy Student\cite{park2020improved} & LS-860 & LSTM & 3.9 & 8.8 & 4.2 & 8.6 \\
        DeCoAR 2.0~\cite{ling2020decoar} & LS-960 & 4-gram & - & - & 5.0 & 12.1 \\
        DiscreteBERT~\cite{baevski2019effectiveness} & LS-960 & 4-gram & 4.0 & 10.9 & 4.5 & 12.1 \\
        wav2vec 2.0 \textsc{Base}~\cite{baevski2020wav2vec}   & LS-960 & 4-gram & 2.7 & 7.9 & 3.4 & 8.0 \\
        wav2vec 2.0 \textsc{Large}~\cite{baevski2020wav2vec}  & LL-60k & Transformer & 1.9 & 4.0 & 2.0 & 4.0 \\
        
        \midrule
        HUBERT \textsc{Base}    & LS-960 & 4-gram & 2.7 & 7.8 & 3.4 & 8.1  \\
        HUBERT \textsc{Large}   & LL-60k & Transformer & 1.8 & 3.7 & 2.1 & 3.9 \\
        HUBERT \textsc{X-Large} & LL-60k & Transformer & 1.7 & 3.0 & 1.9 & 3.5 \\
        
        \bottomrule
    \end{tabular}
    \caption{Results and comparison with the literature on low resource setups (10-min, 1-hour, 10-hour, and 100-hour of labeled data).}
    \label{tab:main_lo}
\end{table*}

\begin{table*}[t]
    \centering
    \begin{tabular}{lcccccc}
        \toprule
        Model & Unlabeled Data & LM & dev-clean & dev-other & test-clean & test-other \\
        \midrule\midrule
        \multicolumn{7}{c}{\textit{\textbf{Superivsed}}} \\
        Conformer L~\cite{gulati2020conformer} & - & LSTM & - & - & 1.9 & 3.9 \\
        
        \midrule\midrule
        \multicolumn{7}{c}{\textit{\textbf{Self-Training}}} \\
        IPL~\cite{xu2020iterative} & LL-60k & 4-gram + Transformer & 1.85 & 3.26 & 2.10 & 4.01 \\
        Noisy Student~\cite{park2020improved} & LV-60k & LSTM & 1.6 & 3.4 & 1.7 & 3.4 \\
        
        \midrule\midrule
        \multicolumn{7}{c}{\textit{\textbf{Pre-Training}}} \\
        wav2vec 2.0 \textsc{Large}~\cite{baevski2020wav2vec} & LL-60k & Transformer & 1.6 & 3.0 & 1.8 & 3.3 \\
        pre-trained Conformer XXL~\cite{zhang2020pushing} & LL-60k & LSTM & 1.5 & 3.0 & 1.5 & 3.1 \\
        
        \midrule\midrule
        \multicolumn{7}{c}{\textit{\textbf{Pre-Training + Self-Training}}} \\
        wav2vec 2.0 + self-training~\cite{xu2020self} & LL-60k & Transformer & 1.1 & 2.7 & 1.5 & 3.1\\
        pre-trained Conformer XXL + Noisy Student~\cite{zhang2020pushing} & LL-60k & LSTM & 1.3 & 2.6 & 1.4 & 2.6 \\
        
        \midrule\midrule
        \multicolumn{7}{c}{\textit{\textbf{This work (Pre-Training)}}} \\
        HUBERT \textsc{Large}   & LL-60k & Transformer & 1.5 & 3.0 & 1.9 & 3.3 \\
        HUBERT \textsc{X-Large} & LL-60k & Transformer & 1.5 & 2.5 & 1.8 & 2.9 \\
        
        \bottomrule
    \end{tabular}
    \caption{Comparison with the literature on high resource setups using all 960 hours of labeled LibriSpeech data.}
    \label{tab:main_hi}
\end{table*}

\section{Results}
\subsection{Main Results: Low- and High-Resource Setups}

Table~\ref{tab:main_lo} presents results for the low-resource setup, where pre-trained models are fine-tuned on 10 minutes, 1 hour, 10 hours, or 100 hours of labeled data. We include comparison with semi-supervised (iterative pseudo labeling (IPL)~\cite{xu2020iterative}, slimIPL~\cite{likhomanenko2020slimipl}, noisy student~\cite{park2020improved}) and self-supervised approaches (DeCoAR 2.0~\cite{ling2020decoar}, DiscreteBERT~\cite{baevski2019effectiveness}, wav2vec 2.0~\cite{baevski2020wav2vec}) in the literature.
Increasing the amount of unlabeled data and increasing the model size improve performance, demonstrating the scalability of the proposed HuBERT self-supervised pre-training method.
In the ultra-low resource setup with just 10 minutes of labeled data, the HuBERT \textsc{Large} model can achieve a WER of 4.7\% on the test-clean set and 7.6\% on the test-other set, which is 0.1\% and 0.6\% WER lower, respectively than the state-of-the-art wav2vec 2.0 \textsc{Large} model. By further scaling up the model size to 1B parameters, the HuBERT \textsc{X-Large} model can further reduce the WER to 4.6\% and 6.8\% on test-clean and test-other. The superiority of HuBERT persists across setups with different amounts of labeled data, with the only exceptions being fine-tuning on 100 hours of labeled data, where HuBERT \textsc{Large} is 0.1\% WER higher than wav2vec 2.0 \textsc{Large} on test-clean, and HuBERT \textsc{Base} is 0.1\% WER higher than wav2vec 2.0 \textsc{Base} on test-other.
In addition, HuBERT also outperforms DiscreteBERT by a large margin in all setups, while both are trained with a virtually identical objective - masked prediction of discovered units. The considerable performance gap suggests two things. First, using waveform as the input to the model is crucial for avoiding loss of information during quantization. Second, while vq-wav2vec~\cite{baevski2019vq}, the units that DiscreteBERT uses for training, may discover better units than k-means clustering of MFCC features, the proposed iterative refinement benefits from the improving HuBERT model and learn better units eventually. We will verify these statements in the ablation study sections.

We report results of fine-tuning HuBERT models on the full 960 hours of Librispeech data and compare with the literature in Table~\ref{tab:main_hi}. Prior studies using additional unpaired speech are classified into:
\begin{enumerate}
    \item self-training: first train an ASR on labeled data to annotate unlabeled speech, and then combine both golden and ASR-annotated text-speech pairs for supervised training.
    \item pre-training: first use unlabeled speech for pre-training a model, and then fine-tune the model on labeled data with a supervised training objective.
    \item pre-training + self-training: first pre-train and fine-tune a model, and then use it to annotate unlabeled speech for self-training combined with supervised data.
\end{enumerate}
HuBERT outperforms the state-of-the-art supervised and self-training methods and is on par with the two best pre-training results in the literature; both are based on wav2vec 2.0 contrastive learning.
In contrast, it lags behind methods combining pre-training with self-training. However, as observed in \cite{xu2020self} and \cite{zhang2020pushing}, we expect that HuBERT can achieve comparable or better performance after combining with self-training, since the pre-trained HuBERT model is on par or better than the pre-trained model those two methods use for pseudo labeling.

\subsection{Analysis: K-Means Stability}
To better understand why masked prediction of discovered units is effective, we conduct a series of analyses and ablation studies. We start with probing the stability of the k-means clustering algorithm concerning different numbers of clusters and different sizes of its training data.
Two features are considered: 39-dimensional MFCC features and 768-dimensional output from the 6-th transformer layer of the first iteration HuBERT-\textsc{Base} model. These two features are used to produce cluster assignments for the first and the second iteration HUBERT training, respectively.

For k-means clustering, we consider $K=\{100,500\}$ clusters fitted on \{1, 10, 100\} hours of speech sampled from the LibriSpeech training split. Each combination of the hyperparameters and the features are trained for 10 trials, and the mean and standard deviation of the supervised PNMI metric on the development set (combining dev-clean and dev-other from LibriSpeech) is reported in Table~\ref{tab:stability}.
The results show that the k-means clustering is reasonably stable given the small standard deviations across different hyperparameters and features. Furthermore, increasing the amount of data used for fitting k-means models improves PNMI in general, but the gain is only as much as 0.012, suggesting the feasibility of using k-means for unit discovery even with limited CPU memory relative to the feature matrix size. Lastly, the PNMI score is much higher when clustering on HuBERT features than clustering on MFCC features, and the gap is even larger with 500 clusters, indicating that iterative refinement significantly improves the clustering quality.

\begin{table}[ht]
    \centering
    \resizebox{\linewidth}{!}{
    \begin{tabular}{cc|ccc}
        \toprule
        \multirow{2}{*}{feature} & \multirow{2}{*}{C} 
        & \multicolumn{3}{c}{PNMI (mean $\pm$ std) with K-means Training Size = } \\
        & & 1h & 10h & 100h \\
        \midrule\midrule
        \multirow{2}{*}{MFCC}
        & 100 & 0.251 $\pm$ 0.001 & 0.253 $\pm$ 0.001 & 0.253 $\pm$ 0.001 \\
        & 500 & 0.283 $\pm$ 0.001 & 0.285 $\pm$ 0.000 & 0.287 $\pm$ 0.001 \\
        \midrule
        \multirow{2}{*}{\textsc{Base}-it1-L6}
        & 100 & 0.563 $\pm$ 0.012 & 0.561 $\pm$ 0.012 & 0.575 $\pm$ 0.008 \\
        & 500 & 0.680 $\pm$ 0.005 & 0.684 $\pm$ 0.003 & 0.686 $\pm$ 0.004 \\
        \bottomrule
    \end{tabular}
    }
    \caption{Stability of K-means as an unsupervised unit discovery algorithm with respect to different features, numbers of clusters, and training data sizes. PNMI stands for phone-normalized mutual information.}
    \label{tab:stability}
\end{table}

\begin{table*}[t]
    \begin{minipage}[c]{0.68\textwidth}
        \centering
        \begin{tabular}{cc|c|ccc}
    \toprule
    \multirow{2}{*}{teacher} & \multirow{2}{*}{C} & 
    \multirow{2}{*}{PNMI} &
    \multicolumn{3}{c}{dev-other WER (\%)} \\
    & & & $\alpha = 1.0$ & $\alpha = 0.5$ & $\alpha = 0.0$ \\
    \midrule\midrule
    Chenone (supervised top-line)  & 8976 & 0.809 & 10.38 & 9.16 & 9.79 \\
    \midrule
    \multirow{3}{*}{K-means on MFCC} 
    & 50  & 0.227 & 18.68 & 31.07 & 94.60 \\
    & 100 & 0.243 & 17.86 & 29.57 & 96.37 \\
    & 500 & 0.276 & 18.40 & 33.42 & 97.66 \\
    \midrule
    K-means on \textsc{Base}-it1-layer6 & 500 & 0.637 & 11.91 & 13.47 & 23.29 \\
    K-means on \textsc{Base}-it2-layer9 & 500 & 0.704 & 10.75 & 11.59 & 13.79 \\
    \bottomrule
\end{tabular}
        \caption{The effect of the training objective and clustering quality on performance. $C$ refers to the number of units, and $\alpha$ is the weight for masked frames.}
        \label{tab:loss}
    \end{minipage}
    \hspace{.5cm}
    \begin{minipage}[c]{0.28\textwidth}
        \centering
        \begin{tabular}{lc}
    \toprule
    teacher & WER \\
    \midrule\midrule
    K-means \{50,100\}       & 17.81 \\
    K-means \{50,100,500\}   & 17.56 \\
    \midrule
    Product K-means-0-100 & 19.26 \\
    Product K-means-1-100 & 17.64 \\
    Product K-means-2-100 & 18.46 \\
    Product K-means-\{0,1,2\}-100 & 16.73 \\
    \bottomrule
\end{tabular}%
        \caption{Cluster ensembles with k-means and product k-means.}
        \label{tab:ens}
    \end{minipage}
\end{table*}

\begin{figure}[h]
  \centering
  \begin{subfigure}[b]{\linewidth}
    \centering
    \begin{tikzpicture}
\begin{axis}[
  title={},
  legend style={font=\tiny},
  ylabel={\small Cluster Purity (\%)},
  ymin=0.0,  ymax=0.4,
  xtick=data,
  legend pos=north west,
  legend columns=3,
  height=5cm,
  width=8cm,
  ymajorgrids=true,
  grid style=dashed,
  ylabel near ticks,
  xlabel near ticks,
]
\addplot[
    color=blue, mark=square, 
] table [y=hyp-pur, x=layer]{figures/qual_it1_km100.tex};
\addlegendentry{C=100, \textsc{Base}-it1}

\addplot[
    color=blue, mark=o, 
] table [y=hyp-pur, x=layer]{figures/qual_it1_km500.tex};
\addlegendentry{C=500, \textsc{Base}-it1}

\addplot[
    color=blue, mark=diamond, 
] table [y=hyp-pur, x=layer]{figures/qual_it1_km1000.tex};
\addlegendentry{C=1000, \textsc{Base}-it1}

\addplot[
    color=red, mark=square, 
] table [y=hyp-pur, x=layer]{figures/qual_it2_km100.tex};
\addlegendentry{C=100, \textsc{Base}-it2}

\addplot[
    color=red, mark=o, 
] table [y=hyp-pur, x=layer]{figures/qual_it2_km500.tex};
\addlegendentry{C=500, \textsc{Base}-it2}

\addplot[
    color=red, mark=diamond, 
] table [y=hyp-pur, x=layer]{figures/qual_it2_km1000.tex};
\addlegendentry{C=1000, \textsc{Base}-it2}

\end{axis}
\end{tikzpicture}
  \end{subfigure}
  \begin{subfigure}[b]{\linewidth}
    \centering
    \begin{tikzpicture}
\begin{axis}[
  title={},
  ylabel={\small Phone Purity (\%)},
  ymin=0.35,  ymax=0.75,
  xtick=data,
  legend style={font=\tiny},
  legend pos=south east,
  legend columns=2,
  height=4cm,
  width=8cm,
  ymajorgrids=true,
  grid style=dashed,
  ylabel near ticks,
  xlabel near ticks,
]
\addplot[
    color=blue, mark=square, 
] table [y=ref-pur, x=layer]{figures/qual_it1_km100.tex};
\addlegendentry{C=100, HUBERT-it1}

\addplot[
    color=red, mark=square, 
] table [y=ref-pur, x=layer]{figures/qual_it2_km100.tex};
\addlegendentry{C=100, HUBERT-it2}

\addplot[
    color=blue, mark=o, 
] table [y=ref-pur, x=layer]{figures/qual_it1_km500.tex};
\addlegendentry{C=500, HUBERT-it1}

\addplot[
    color=red, mark=o, 
] table [y=ref-pur, x=layer]{figures/qual_it2_km500.tex};
\addlegendentry{C=500, HUBERT-it2}

\addplot[
    color=blue, mark=diamond, 
] table [y=ref-pur, x=layer]{figures/qual_it1_km1000.tex};
\addlegendentry{C=1000, HUBERT-it1}

\addplot[
    color=red, mark=diamond, 
] table [y=ref-pur, x=layer]{figures/qual_it2_km1000.tex};
\addlegendentry{C=1000, HUBERT-it2}

\legend{}
\end{axis}
\end{tikzpicture}
  \end{subfigure}
  \begin{subfigure}[b]{\linewidth}
    \centering
    \begin{tikzpicture}
\begin{axis}[
  title={},
  legend style={font=\tiny},
  xlabel={\small Layer},
  ylabel={\small PNMI (\%)},
  ymin=0.35,  ymax=0.75,
  xtick=data,
  legend pos=south east,
  legend columns=2,
  height=4cm,
  width=8cm,
  ymajorgrids=true,
  grid style=dashed,
  ylabel near ticks,
  xlabel near ticks,
]
\addplot[
    color=blue, mark=square, 
] table [y=MI/H(ref), x=layer]{figures/qual_it1_km100.tex};
\addlegendentry{C=100, HUBERT-it1}

\addplot[
    color=red, mark=square, 
] table [y=MI/H(ref), x=layer]{figures/qual_it2_km100.tex};
\addlegendentry{C=100, HUBERT-it2}

\addplot[
    color=blue, mark=o, 
] table [y=MI/H(ref), x=layer]{figures/qual_it1_km500.tex};
\addlegendentry{C=500, HUBERT-it1}

\addplot[
    color=red, mark=o, 
] table [y=MI/H(ref), x=layer]{figures/qual_it2_km500.tex};
\addlegendentry{C=500, HUBERT-it2}

\addplot[
    color=blue, mark=diamond, 
] table [y=MI/H(ref), x=layer]{figures/qual_it1_km1000.tex};
\addlegendentry{C=1000, HUBERT-it1}

\addplot[
    color=red, mark=diamond, 
] table [y=MI/H(ref), x=layer]{figures/qual_it2_km1000.tex};
\addlegendentry{C=1000, HUBERT-it2}

\legend{}
\end{axis}
\end{tikzpicture}
  \end{subfigure}
  \caption{Quality of the cluster assignments obtained by running k-means clustering on features extracted from each transformer layer of the first and the second iteration \textsc{Base} HuBERT models.}
  \label{fig:qual_layer}
\end{figure}
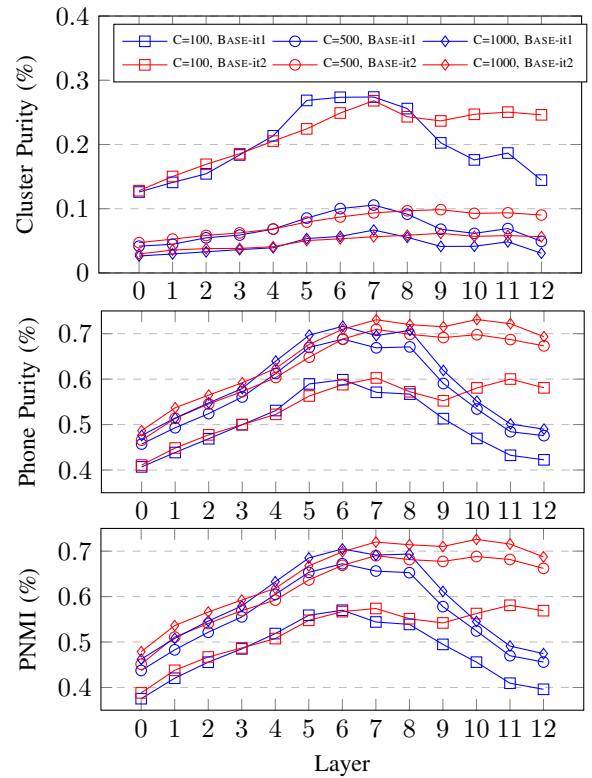

\subsection{Analysis: Clustering Quality Across Layers and Iterations}
We next study how each layer of the HuBERT model from each iteration performs when used for clustering to generate training targets.
The two \textsc{Base} HuBERT models from the first two iterations as described in Section~\ref{sec:pretrain} are considered, which are referred to as \textsc{Base}-it1 and \textsc{Base}-it2, respectively. There are 26 features representing 12 transformer layers plus the input to the first transformer layer (denoted as ``Layer 0'') from the two HuBERT models.
For each feature, we fit three k-means models ($K=\{100, 500, 1000\}$ clusters) on a 100 hour subset randomly sampled from the LibriSpeech training data. The teacher quality measured in cluster purity, phone purity, and phone normalized mutual information (PNMI) is shown in Figure~\ref{fig:qual_layer}.
As a baseline, MFCC achieves (cluster purity, phone purity, PNMI) = (0.099, 0.335, 0.255) for $K=100$ and (0.031, 0.356, 0.287) for $K=500$.

Both \textsc{Base}-it1 and \textsc{Base}-it2 features result in significantly better clustering quality on all three metrics than MFCC with the same number of clusters. On the other hand, the best \textsc{Base}-it2 feature is better than the best \textsc{Base}-it1 on phone purity and PNMI, but slightly worse on cluster purity.
Finally, we observe different trends across layers from \textsc{Base}-it1 and \textsc{Base}-it2: while \textsc{Base}-it2 model features generally improve over layers, \textsc{Base}-it1 has the best features in the middle layers around the 6th layer. Interestingly, the quality of the last few layers degrades dramatically for \textsc{Base}-it1, potentially because it is trained on target assignments of worse quality, and therefore the last few layers learn to mimic their bad label behavior.

\subsection{Ablation: The Importance of Predicting Masked Frames}
We present a series of ablation studies in the following sections to learn how pre-training objective, cluster quality, and hyperparameters affect the performance. 
The models for ablation studies are pre-trained for 100k steps and fine-tuned on the 10-hour libri-light split using fixed hyperaprameters. MFCC-based k-means units with C=100 are used if not otherwise mentioned. We report WERs on the dev-other set decoded with the $n$-gram language model using fixed decoding hyperparameters.

To understand the importance of our proposal to predict the masked frames only, we compare three conditions: 1) predicting masked frames, 2) predicting all frames, and 3) predicting unmasked frames, which can be simulated by setting $\alpha$ to 1.0, 0.5, and 0.0, respectively. 
We are comparing three k-means models learned from clustering MFCC teachers with 50, 100, 500 clusters, one learned from clustering HuBERT-\textsc{Base}-it1 6th transformer layer features, and supervised labels obtained from the forced-alignment of character-based HMM models (chenone)~\cite{le2019senones}.

Results shown in Table~\ref{tab:loss} indicate that when learning from bad cluster assignments, computing loss only from the masked regions achieves the best performance, while the inclusion of unmasked loss results in significantly higher WERs. 
However, as the clustering quality improves, the model would suffer less when computing losses on the unmasked frames (\textsc{Base}-it1-layer6) or even achieve better performance as the case of chenone.

\subsection{Ablation: The Effect of Cluster Ensembles}
To understand the effect of combining multiple k-means models for generating targets, we consider two setups. The first one has k-means models of different numbers of clusters presented in Table~\ref{tab:loss}, denoted with KM-\{50,100,500\}. The second one has k-means models trained on spliced MFCC features with a window of three; hence, each input feature is represented as a 117-dimensional vector. In this second case, we apply product quantization on the spliced features, where dimensions are split into the coefficients of the zeroth, first, and second-order derivatives, with each 39-dimensional subspace quantized to a codebook of 100 entries. We denote these codebooks with Product k-means-\{0,1,2\}-100, respectively.
By comparing the results from Table~\ref{tab:loss} and Table~\ref{tab:ens}, it is clear that using an ensemble leads to better performance than what a single k-means clustering can achieve.

\subsection{Ablation: Impact of Hyperparameters}
\begin{figure}[ht]
  \centering
  \begin{subfigure}[b]{0.49\linewidth}
    \centering
    \begin{tikzpicture}
\begin{axis}[
  xlabel={\small $p$},
  ylabel={\small WER (\%)},
  ymin=16,
  ymax=24,
  xtick=data,
  width=\linewidth,
  height=2.5cm,
  ymajorgrids=true,
  grid style=dashed,
  ylabel near ticks,
  xlabel near ticks,
  ticklabel style={font=\scriptsize},
]
\addplot table [y=km100_100k, x=maskp]{figures/maskp.tex};
\end{axis}
\end{tikzpicture}
  \end{subfigure}
  \begin{subfigure}[b]{0.49\linewidth}
    \centering
    \begin{tikzpicture}
\begin{axis}[
  xlabel={\small \#GPUs},
  ylabel={\small WER (\%)},
  ymin=15,
  ymax=40,
  xtick=data,
  width=\linewidth,
  height=2.5cm,
  ymajorgrids=true,
  grid style=dashed,
  ylabel near ticks,
  xlabel near ticks,
  ticklabel style={font=\scriptsize},
]
\addplot table [y=km100_100k, x=ngpu]{figures/batchsize.tex};
\end{axis}
\end{tikzpicture}
  \end{subfigure}
  \caption{Varying masking probability $p$ (left) and effective batch size through the number of GPUs (right).}
  \label{fig:prob_bs}
\end{figure}
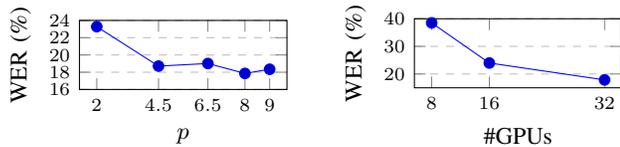

Figure~\ref{fig:prob_bs} and Table~\ref{tab:step} studies how hyperparameters affect HuBERT pre-training.
It is shown that
\begin{enumerate*}[label=(\arabic*)]
    \item the portion of frames selected as mask start is optimal at $p=$8\%;
    \item increasing the batch size can significantly improve the performance; 
    \item training for longer consistently helps for both k-means models with C=\{50, 100\}, and the best model achieves a WER of 11.68\%.
\end{enumerate*}
These findings are also consistent with those from BERT-like models~\cite{clark2020electra}. In addition, we include a comparable result from DiscreteBERT~\cite{baevski2019effectiveness} in Table~\ref{tab:step} which applies k-means to quantize the same MFCC features into 13.5k units, used as both the output and the \textit{input} to the BERT model. Besides using continuous speech input rather than discrete units, We hypothesize that HuBERT achieves significantly better performance because its fewer k-means clusters of 100 or 500 help capture broad phonetic concepts without delving into inter/intra-speaker variation. 

\begin{table}[ht]
    \centering
    \begin{tabular}{cc|cccc}
    \toprule
    \multirow{2}{*}{teacher} & \multirow{2}{*}{C} &  
    \multicolumn{4}{c}{dev-other WER (\%)} \\
    & & steps=100k & 250k & 400k & 800k \\
    \midrule
    \multirow{2}{*}{K-means}  
    & 50  & 18.68 & 13.65 & 12.40 & 11.82 \\
    & 100 & 17.86 & 12.97 & 12.32 & 11.68 \\
    \midrule
    \cite{baevski2019effectiveness} & 13.5k & \multicolumn{4}{c}{26.6} \\
    \bottomrule
\end{tabular}
    \caption{Varying the number of HuBERT pre-training steps. $p$ is set to 6.5\%.}
    \label{tab:step}
\end{table}

\section{Conclusion}
This paper presents HuBERT, a speech representation learning approach that relies on predicting K-means cluster assignments of masked segments of continuous input. On both the Librispeech 960 hours and the 60,000 hours Libri-light pre-training setups, HuBERT matches or outperforms the state-of-the-art systems over all fine-tuning subsets of 10mins, 1h, 10h, 100h, and 960h. Furthermore, the learned representation quality improves dramatically with iteratively refining K-means cluster assignments using learned latent representations for a previous iteration. Finally, HuBERT scales well to a 1B transformer model showing a relative reduction in WER of up to 13\% on the test-other subset. For future work, we plan to improve the HuBERT training procedure to consist of a single phase. Furthermore, given the high quality of its representations, we will consider using HuBERT pre-trained representations for multiple downstream recognition and generation tasks beyond ASR.

\ifCLASSOPTIONcaptionsoff
  \newpage
\fi

\bibliographystyle{IEEEtran}
\bibliography{refs}

\end{document}